\documentclass[letterpaper, 10 pt, conference]{ieeeconf}  

\IEEEoverridecommandlockouts                              
\usepackage{amsmath,amssymb,amsfonts}
\usepackage{graphics} 

\usepackage{xspace}
\PassOptionsToPackage{usenames,dvipsnames}{xcolor}
\usepackage[hidelinks]{hyperref}
\usepackage{gensymb}
\usepackage[noend]{algpseudocode}
\usepackage{multirow}
\usepackage{algorithm}
 \usepackage{float}
\usepackage{subcaption}
\usepackage{stfloats}
\usepackage{graphicx}
\usepackage{wasysym}
\usepackage{tcolorbox}
\usepackage{booktabs} 
\usepackage{siunitx}
\usepackage[percent]{overpic}
\usepackage{textcomp}
\usepackage{tikz}
\newcommand{\calA}{\ensuremath{\mathcal{A}}\xspace}

\newcommand{\calQ}{\ensuremath{\mathcal{Q}}\xspace}

\newcommand{\calW}{\ensuremath{\mathcal{W}}\xspace}

\newcommand{\calO}{\ensuremath{\mathcal{O}}\xspace}

\newcommand{\R}{\mathbb{R}}

\newcommand{\N}{\mathbb{N}}

\newtheorem{prob}{Problem}

\newcommand{\ignore}[1]{}

\newcommand{\Cpp}{C\raise.08ex\hbox{\tt ++}\xspace}
\newcommand{\open}{{\sc Open}\xspace}
\newcommand{\closed}{{\sc Closed}\xspace}

\newcommand{\stepone}{\texttt{PopulateCCArcs(pose)}\xspace}
\newcommand{\steptwo}{\texttt{IdenitfyContact(cell)}\xspace}
\newcommand{\mbf}{\mathbf}
\newcommand{\figref}[1]{\mbox{Fig.~\ref{#1}}}

\newcommand{\la}{$\leftarrow$}

\newcommand{\boldk}{\mathbf{\kappa}}
\newcommand{\angleX}{\psi}

\newcommand{\lseg}{\ell_{\mathrm{seg}}}
\newcommand{\lten}{\ell_{\mathrm{ten}}}
\newcommand{\qinit}{\mbf{q}_{\mrm{init}}}
\newcommand{\ee}{end-effector\xspace}
\newcommand{\mrm}{\mathrm}

\newcommand{\config}{\mathbf{c}(\mathbf{q})}

\newcommand{\bolddelta}{\mathbf{\delta}}
\newcommand{\pose}[1]{\left(\mbf{p}\textsubscript{#1}, \angleX_{#1}\right)}
\newcommand{\position}{\mbf{u}}
\newcommand{\steponeinput}[1]{\texttt{PopulateCCArcs#1}}

\newcommand{\mcan}{$\mathcal{M}_{\text{CAN}}$}

\newcommand\copyrighttext{%
    \footnotesize \color{gray}
    \textcopyright  
    2024
    \ IEEE. Personal use of this material is permitted.  Permission from IEEE must be obtained for all other uses, in any current or future media, including reprinting/republishing this material for advertising or promotional purposes, creating new collective works, for resale or redistribution to servers or lists, or reuse of any copyrighted component of this work in other works.
}

\title{\LARGE \bf
Towards Contact-Aided Motion Planning\\for Tendon-Driven Continuum Robots
}
\author{Priyanka Rao$^{1}$,~\IEEEmembership{Student~Member,~IEEE}, Oren Salzman$^{2}$,~\IEEEmembership{Member,~IEEE},\\
and Jessica Burgner-Kahrs$^{1}$,~\IEEEmembership{Senior~Member,~IEEE}
\thanks{%
$^{1}$Continuum Robotics Laboratory, Department of Mathematical \& Computational Sciences, University of Toronto, ON, Canada
}%
\thanks{%
$^{2}$Department of Computer Science, Technion, Haifa
3200003, Israel
}%
\thanks{%
{\tt\small priyankaprakash.rao@mail.utoronto.ca.
 }
 }%
}

\begin{document}
\maketitle
\thispagestyle{empty}
\pagestyle{empty}

\begin{abstract}
Tendon-driven continuum robots (TDCRs), with their flexible backbones, offer the advantage of being used for navigating complex, cluttered environments. However, to do so, they typically require multiple segments, often leading to complex actuation and control challenges.
To this end, we propose a novel approach to navigate cluttered spaces effectively for a single-segment long TDCR which is the simplest topology from a mechanical point of view.
Our key insight is that by leveraging contact with the environment we can achieve multiple curvatures without mechanical alterations to the robot.
Specifically, we propose a search-based motion planner for a single-segment TDCR. This planner, guided by a specially designed heuristic, discretizes the configuration space and employs a best-first search. The heuristic, crucial for efficient navigation, provides an effective cost-to-go estimation while respecting the kinematic constraints of the TDCR and environmental interactions. 
We empirically demonstrate the efficiency of our planner---testing over 525 queries in environments with both convex and non-convex obstacles,
our planner is demonstrated to have a success rate of about~80\% 
while baselines were not able to obtain a success rate higher than~30\%. 
The difference is attributed to our novel heuristic which is shown to significantly reduce the required search space. 
\end{abstract}
\begin{tikzpicture}[remember picture,overlay]
        \node[anchor=south,yshift=10pt] at (current page.south) {\parbox{\dimexpr0.75\textwidth-\fboxsep-\fboxrule\relax}{\copyrighttext}};
\end{tikzpicture}
\section{Introduction}

Tendon-driven continuum robots (TDCRs), bend their flexible backbones via tendon actuation, making them popular for medical and industrial applications for accessing areas with constricted entry-points. 
As navigating cluttered areas requires more intricate shapes, several segments are typically used, as each segment is capable of bending only in a C-shape. A multi-segment design comes at the price of complex actuation unit design \cite{russo2023} and tendon control.
While methods like variable tendon routing \cite{Gao2015}, layer jamming \cite{Kim2012}, and locking mechanisms \cite{Pogue2022} vary curvatures, they demand mechanical modifications to the robot design. 
In this work we argue that in cluttered environments, we can use a simple single-segment long TDCR that exploits contacts with the environment (\emph{contact-aided navigation (CAN)}) to achieve variations in curvature. 

TDCRs  can be used to deviate from the common obstacle-avoidance paradigm, as they can interact with the environment due to their compliance. For example, they have been exploited for environmental sensing by actively tapping the surroundings \cite{Wooten2022}. 
They can be used for CAN as the environmental interactions deform the robot due to resulting external constraints on the robot shape, providing passive degrees of freedom (DOF) that can be leveraged to improve the  workspace that the robot can access~\cite{Selvaggio2020}.  However, in scenarios cluttered with obstacles where the robot's end-effector must reach a target pose in order to carry out various tasks, a vital component is a motion planner capable of making combinatorial decisions about the necessary contacts to reach the desired pose. This motion planner, which is the focus of this paper, is responsible for identifying the sequence of actuation inputs required to navigate the robot.

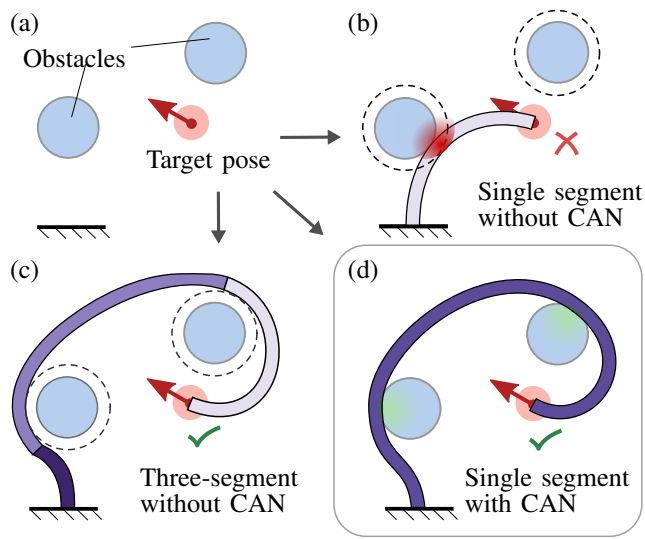
\begin{figure}[t!]
    \begin{overpic}[width=\columnwidth, grid = False]{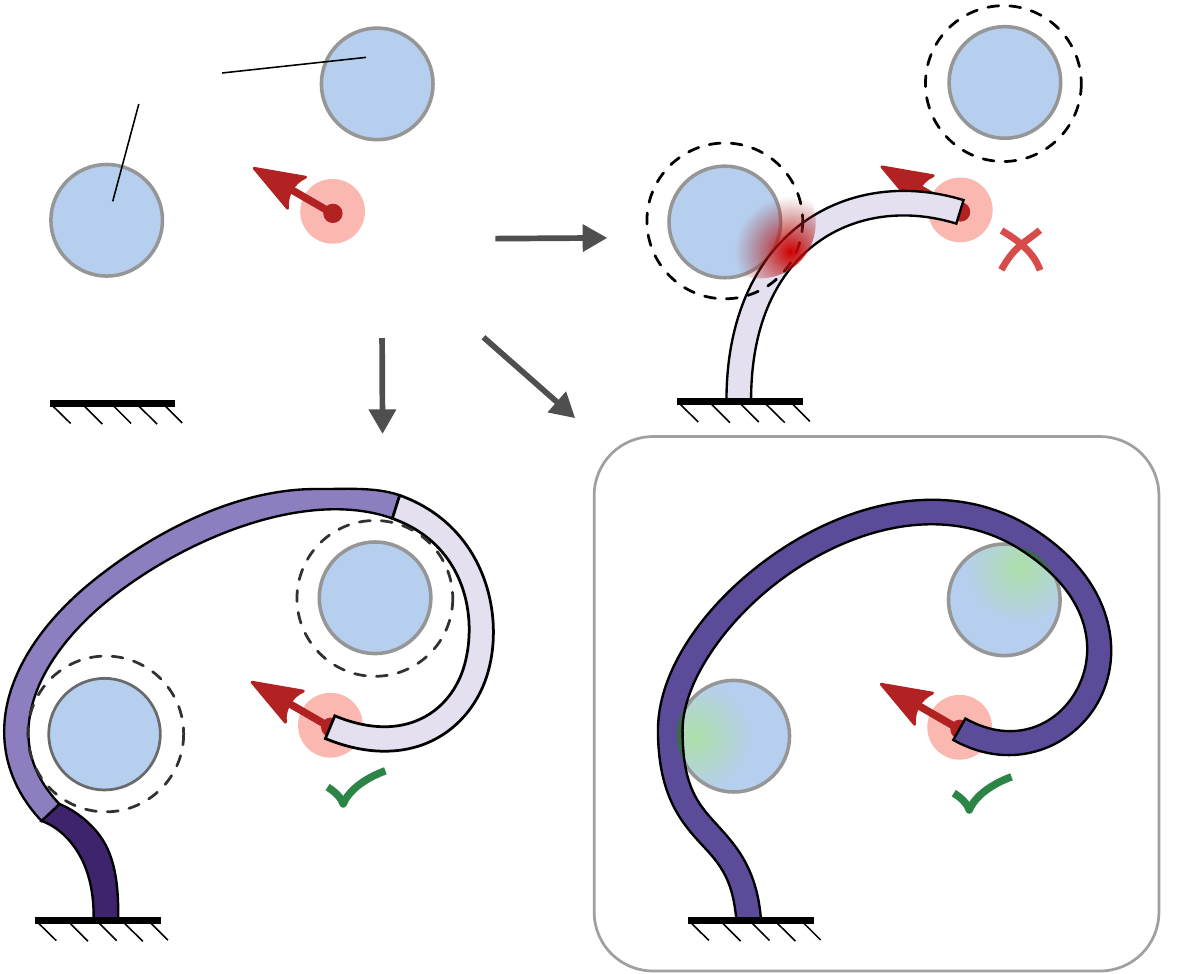}
    \put(2,72.5){Obstacles}
    \put(21,57){Target pose}
    \put(0,78){(a)}
    \put(52,78){(b)}
    \put(0,40){(c)}
    \put(52,40){(d)}
    \put(70,8){Single segment}
    \put(70,4){with CAN}
    \put(72,52){Single segment}
    \put(72,48){without CAN}
    \put(20,8){Three-segment}
    \put(20,4){without CAN}
    \end{overpic}
    \caption{(a) Problem statement: the end-effector of a TDCR must reach a  given target pose in a cluttered environment, 
    (b) Single-segment TCDR's  unsuccessful solution (without CAN) which intersects an obstacle, 
    (c) Three-segment TDCR's successful solution, 
    (d) Single-segment's successful solution (with CAN) obtained by leveraging the obstacles to vary its curvature and reach the target pose.}
     \label{fig:motivation}
     \vspace{-1em}
\end{figure}

Consider an environment with obstacles, where the robot needs to reach a target pose (see \figref{fig:motivation}(a)). A single-segment robot has limited workspace and cannot achieve the illustrated target pose, while avoiding contact with obstacles (\figref{fig:motivation}(b)). To reach the target pose with obstacle avoidance, a multi segment robot is potentially required (\figref{fig:motivation}(c)). A single segment long TDCR is essentially under-actuated for the task at hand as it only has two actuated DOFs (from bending and length insertion). However, CAN allows it to potentially leverage contacts with the obstacles to vary its curvature and reach the target pose, as shown in \figref{fig:motivation}(d).

\subsection{Related Work}
Soft growing vine robots, despite being underactuated, leverage their follow-the-leader (FTL) deployment capability for CAN. However, the robot path is modeled as a series of line segments \cite{Greer2020,Fuentes2023} to simplify trajectory generation between obstacles, and does not reflect the TDCR's continuous backbone.
Selvaggio et al. \cite{Selvaggio2020} use inverse kinematics to actively steering vine robots, but only look at point-sized obstacles. 

CAN has been exploited for hyperredundant robots where they have the advantage of additional DOFs that can be used to steer the end-effector towards the target pose.  

Sampling-based approaches like Rapidly-exploring Random Trees \cite{PFOTZER2017123} can perform poorly when required to traverse through narrow passages present between obstacles \cite{Saleem2021}.

Search-based planning in the configuration space has been proposed to reach a target position \cite{Singh2018}, a target pose \cite{Saleem2021} and to find stable load configurations \cite{Wang2021} for these multi-link robots. 
While search-based planning offers a more systematic approach to exploring the workspace, the proposed heuristics in the above works are tailored for multi-link robots, which offer increased maneuverability (due to higher DOF) for controlling the end-effector's pose.

Therefore, CAN for a single segment TDCR necessitates methods that respect the robot kinematics and the fact that it is underactuated.
Existing controllers account for possible contacts \cite{Yip2014, Zhang2019}, but require user-defined trajectories.
Given a trajectory, inverse-kinematics models respecting contact mechanics \cite{coevoet2017} could be used to follow the trajectory as well.

Naughton et al.~\cite{Naughton2021} use reinforcement learning to develop the motion plan for a two-segment soft robot for position control amidst obstacles, although their work does not extend to achieving a target pose. 

\subsection{Contributions}
In this letter, we propose a search-based motion planner for a planar single-segment TDCR to reach a target position and orientation. 
Our search-based planner discretizes the configuration space, and performs a best-first search  while exploiting contacts with the environment. 
Key to the efficiency of our planner is a novel efficient-to-compute heuristic function. It captures the kinematics of the single segment TDCR and guides the planner to the goal, while exploring only a fraction of the entire search space.

\section{Problem Definition}
We consider a planar single-segment TDCR and denote its~2D workspace as $\calW\in \mathbb{R}^2$, consisting of $o$ known obstacles $\calW_{\mathrm{obs}} = \{\calO_1, \ldots \calO_o\}$. The boundary of the $j^{\text{th}}$  obstacle is defined by a closed curve $f_j(\mbf{x})=0$ with points lying outside the obstacle satisfying $f_j(\mbf{x})>0$. 

Our robot is actuated via 
(i)~inserting and retracting the base of the robot
and
(ii)~pulling and releasing a single tendon.\footnote{ 
The bending in a TDCR is underactuated, where even though there are two antagonistic tendons, and thus two inputs, the robot only has one degree of freedom in bending from tendon actuation. For simplification, we consider the length of one tendon, $\lten^{1}$ as the actuation input. Once the corresponding robot configuration is calculated, the value of the second tendon, $\lten^{,2}$ can be determined using geometry.
We remove the superscript of the first tendon for convenience.
}

To this end, a \emph{joint-space value},~$\mbf{q} = (\lseg, \lten)$ describes the total length of the robot that has been inserted into the workspace and the tendon length, respectively.
We denote the joint-space, namely the space of all joint-space values, as~$\calQ$.
A \emph{joint-space action}~$\bolddelta = (\delta_{\mathrm{seg}} ,\delta_{\mathrm{ten}})$ describes 
the amount~$\delta_{\mathrm{seg}}$ that the base is inserted or retracted (corresponding to positive and negative values of $\delta_{\mathrm{seg}}$, respectively) 
and
the amount~$\delta_{\mathrm{ten}}$ that the tendon is releasaed or pulled (corresponding to positive and negative values of $\delta_{\mathrm{ten}}$, respectively). 
Applying a joint-space action~$\bolddelta = (\delta_{\mathrm{seg}} ,\delta_{\mathrm{ten}})$
on  
a robot with the joint-space value~$\mbf{q} = (\lseg, \lten)$
is denoted by $\mbf{q} + \bolddelta$ 
and yields the new joint-space value $( \lseg + \delta_{\mathrm{seg}},\lten + \delta_{\mathrm{ten}})$.
We will use  $\sigma$ to denote a \emph{sequence} of joint-space actions and $\Sigma$ to denote all such possible sequences.

Applying a joint-space action on a joint-space value is not always guaranteed to result in a feasible solution as the model might return a solution that does not satisfy the problem constraints.
We refer to such actions as \emph{invalid actions}, and all actions leading to feasible solutions as \emph{valid actions}.
Finally, we denote by $\mbf{p}({\mbf{q} + \sigma})$ and $\angleX({\mbf{q} + \sigma})$ the position and orientation of the robot's end-effector w.r.t. to the global frame of reference after applying $\sigma$ to~$\mbf{q}$.

Now, the motion planning problem can be stated as follows.

\begin{prob}
\label{prob:mp}
Given 
a single-segment TDCR with known shape known for an initial joint-space value $\qinit \in Q$,
moving amidst a set $\calO = \{\calO_1, \ldots \calO_o\}$ of known obstacles,
a tip goal pose, $\mbf{T}_{\mrm{goal}}$ containing the goal position and orientation~$\mbf{p}_{\mrm{goal}},\mbf{\angleX}_{\mrm{goal}}$,
a position and orientation  tolerance $\varepsilon,\omega$,
our motion-planning problem calls for computing a sequence of valid joint space actions $\sigma = \{\bolddelta_1, \ldots, \bolddelta_k \}$  such that
\begin{align}
    \vert\vert\mbf{p}({\qinit + \sigma}) - \mbf{p}_{\mrm{goal}} \vert\vert_2 \leq \varepsilon,\\
    \vert\mbf{\angleX}({\qinit + \sigma}) - {\angleX}_{\mrm{goal}} \vert \leq \omega.
\end{align}
Namely, that the robot's tip reaches the goal pose within the goal tolerance as shown in \figref{fig:kinematic}(a).
And that each joint-space action $\bolddelta_i$ is a valid action when applied to $\qinit + \bolddelta_1 + \ldots \bolddelta_{i-1}$.

\end{prob}

\section{Kinematic model for obstacle interaction}
\label{section:model}
\begin{figure}[t]
    \centering
    \begin{overpic}[width=1\columnwidth,tics=5, grid = False]{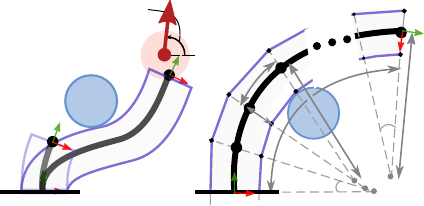}
        \put(77,20){\rotatebox{-60}{{$1/\kappa_3$}}}
        \put(95,25){\rotatebox{-90}{{$1/\kappa_m$}}}
        \put(61,24){\rotatebox{-35}{{$r_d$}}}
        \put(89,25){{$\theta_m$}}
        \put(56,4){{$x$}}
        \put(52,7){{$y$}}
         \put(74,4){{$\theta_1$}}
         \put(52,30){\rotatebox{41}{{$\lseg/m$}}}
         \put(76,30){\rotatebox{30}{{$\lten$}}}
        \put(25,46){$\angleX_{\mrm{goal}}$}
        \put(0,17){{$\mbf{p}(\qinit)$}}
        \put(15,31){{$\mbf{p}(\qinit+\sigma)$}}

        \put(28,38){$\mbf{p}_{\mrm{goal}}$}
        \put(30,0){(a)}
        \put(90,0){(b)}
    \end{overpic} 
    \caption{(a)~Robot shape at $\qinit$ and at $\qinit+\sigma$ with respective end-effector positions.
    (b)~Workspace $\calW$ with obstacles indicated in blue. 
    The backbone is represented by a sequence of constant-curvature arcs subtending an angle~$\theta_i$ with curvatures~$\kappa_i$, where $i=1,\ldots,m$. The tendons are indicated in violet. 
    }
    \label{fig:kinematic}
\end{figure}
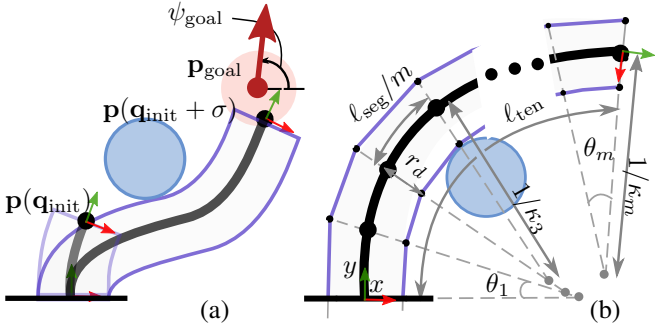

In this section we describe our kinematic model for a planar single-segment TDCR.
Our approach adapts the model introduced by Ashwin et al.~\cite{Ashwin2021}.\footnote{This model assumes there is no friction in the system and that the tendons are partially constrained \cite{Rao2021survey}.}
However, to reduce computational complexity, instead of employing the proposed four-bar mechanism, we use the so-called \emph{piece-wise constant-curvature assumption} (PCCA). PCCA discretizes the robot into a series of $m$ mutually-tangent constant-curvature arcs, each of curvature $\kappa_i,\forall i=1,2\ldots m$ as shown in \figref{fig:kinematic}(b). 
These individual curvatures are collectively represented by the vector $\boldk$. The \emph{configuration-space value} is represented by $\config=(\lseg, \boldk)$, 
and can be used to reconstruct the robot shape. As we will see, the kinematic model is treated as an energy-minimization problem, where the equilibrium configurations must
(i)~satisfy input tendon length constraints and
(ii)~remain outside the obstacles.
In this section we first describe the optimization based kinematic model, and then use it to describe the required backbone parameters.

\subsection{Optimization-based kinematic model}

The objective of the kinematic model is to determine the values of $\boldk$ for a given joint-space value $\mbf{q}$, while ensuring that the robot lies outside the obstacles. 
Since we consider only one tendon as the actuation input, the first nonlinear equality constraint enforces that the actuated tendon length,~$\lten$ should be equal to the calculated tendon length~$\lten^*$. Calculating these tendon lengths for a configuration $\mbf{c}$ is described in Sec.~\ref{subsec:kinematic-rep}. 
\begin{align}
    g(\config)=\lten^*-\lten.
    \label{eqn:equality_constraint}
\end{align}

Recall that  the boundary of an obstacle $j$ is defined by the function ${f}_j(\mbf{x})$. Let  angle $\angleX_{f_j}(\mbf{x})$ be the angle made by the tangent to this obstacle at $\mbf{x}$. Since we want the robot to lie outside these obstacles, we can formulate the nonlinear constraint as 
\begin{align}
    {f}_j(\mbf{X}(\config))\geq0,
    \label{eqn:inequality constraints}
\end{align}
where $\mbf{X(\config)}$ represents a set of points approximating the robot's geometry. The selection of these points is detailed in Sec.~\ref{subsec:kinematic-rep}. 
Assuming the material is isotropic, minimizing the strain energy at equilibrium configurations is equivalent to minimizing the individual curvatures of the subsegments~\cite{Ashwin2021}. The resulting optimisation problem, using Eq.~\eqref{eqn:equality_constraint} and \eqref{eqn:inequality constraints}, is given by
\begin{align}
    \min_{\boldk} \quad & \sum_{i=1}^n(\kappa_i)^2\nonumber\\
\textrm{s.t.} \quad & g(\config)=0\nonumber\\
  &  {f}_j(\mbf{X}(\config))\geq0, \forall j=1,2 \ldots o.
  \label{eqn:optimisation_problem}
\end{align}

\subsection{Robot kinematic representation}
\label{subsec:kinematic-rep}
Recall that we discretize the backbone into a series of~$m$ circular arcs, also called \emph{subsegments}\footnote{Such discretization is natural as typically, a TCDR is built using a series of $m$ disks used to route the tendons.}. For a robot of length~$\lseg$, each sub-segment has a length of $\lseg / m$ and curvature of $\kappa_i$. Thus, the subtending angle of the $i$'th sub-segment is $\theta_i=\kappa_i \cdot \lseg/m$. The vector $\boldk$ contains the curvatures of all these subsegments. 

The tip of each subsegment is denoted by $\mbf{p}_i$. If we define a local frame of reference attached to $p_i$ with its $z$-axis tangent to the backbone, 
then the transformation matrix $\mbf{T}_{i-1}^{i}\in \mathrm{SE}(2)$ is used to calculate the relative pose between two such consecutive frames.
We consider the two tendons  placed at a distance of $r_d$ from the backbone and  denote their  position vectors at the $i$'th sub-segment as $\mbf{p}^{1}_{i,\rm{ten}}$ and $\mbf{p}^{2}_{i,\rm{ten}}$ respectively.\footnote{See \cite{Rao2021survey} for the exact formulae.}
Eq.~\eqref{eqn:equality_constraint} requires calculating the length of the first tendon which is given by
\begin{align}
    \lten^{}(\config)=\sum_{i=1}^{m}||\mbf{p}^1_{i,\rm{ten}}-\mbf{p}^1_{i-1,\rm{ten}}||_2.
    \label{eqn:tendon_length}
\end{align}

To define the nonlinear constraints in Eq.~\eqref{eqn:inequality constraints}, we need to approximate the robot shape by a set of discrete points. In our setting, we will use a set of backbone and tendon positions. Specifically for a given $\config$, we define the set $\mbf{X}(\mathbf{q}, \boldk)$ of~$3m$ points as
\begin{align}
    \mbf{X(\config)}=\bigg\{\begin{bmatrix}
        \mbf{p}_i &
        \mbf{p}^1_{i,\rm{ten}} &
        \mbf{p}^2_{i,\rm{ten}}
    \end{bmatrix}, \forall i=1,\ldots, m\bigg\}.
    \label{eqn:shape_X}
\end{align}

\subsection{Applying a sequence of joint space actions}

\begin{figure}[t!]
    \begin{overpic}[width=\columnwidth]{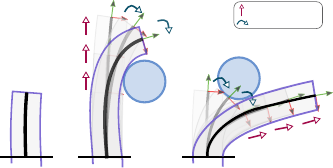}
    \put(22,24){1}
    \put(22,34){2}
    \put(22,42){3}
    \put(44,46){4}
    \put(53,41){5}
    \put(64,29){1}
    \put(74,25){2}
    \put(78,12){3}
    \put(86,14){4}
    \put(92,16){5}
    \put(13,4.5){(a) $\mbf{q}$}
    \put(37,4.5){(b) $\mbf{q}+\sigma$}
    \put(69,4.5){(c) $\mbf{q}+\sigma'$}
    \put(75,47){\scriptsize{Length extension}}
    \put(75,43){\scriptsize{Bending}}
    \end{overpic}
    \caption{(a) Robot's initial configuration at $\mbf{q}$. (b), (c) Overlayed robot shapes for joint sequences (b) $\sigma$ (two length-extensions, and three bending actions) and (c) $\sigma'$ (reverse order: three bending actions, and two length-extensions).}
    \label{fig:redundancy}
\end{figure}
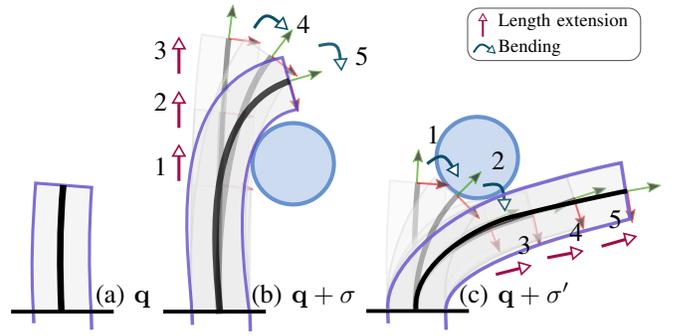

In the absence of obstacles, there is a one-to-one mapping between a sequence of joint-space actions applied to the robot starting at some initial joint-space value and the final shape of the robot.
However, contacts between the robot and the obstacles affect its shape.
This implies that given 
an initial joint-space value~$\mbf{q} = (\lseg, \lten)$
and 
two different joint-space sequences $\sigma, \sigma'$ such that $\mbf{q} + \sigma = \mbf{q} + \sigma'$,
may result in different robot shapes due to different contact histories. An example is illustrated in \figref{fig:redundancy}.
Therefore, there can be multiple solutions for Eq.~\eqref{eqn:optimisation_problem}.  

To solve Problem \ref{prob:mp}, we begin with the initial joint-space value $\qinit$, with known $(\ell_{\mrm{init}}, \boldk_{\mrm{init}})$. To compute the resulting shape after applying action $\bolddelta_1$ to $\qinit$, we use $\boldk_{\mrm{init}}$ to warm-start the optimization in Eq.~\eqref{eqn:optimisation_problem}. The obtained curvature values can be used to calculate the end-effector position and orientation, and then used to solve for $(\qinit+\bolddelta_1+\bolddelta_2)$. The model therefore, needs to be solved~$k$ times to obtain the $\boldk$ for $(\qinit+\bolddelta_1+\bolddelta_2\ldots +\bolddelta_k)$. Subsequently, the end-effector's position and orientation are determined.

\section{Method}
To address our motion-planning problem, we take a search-based planning approach~\cite{CohenPL14,FuSA21} where we run a best-first search on a discretized approximation of the continuous joint-space.
This section details our implementation of the necessary primitive operations, data structures, and the proposed heuristic for this search.

\subsection{Node representation}
\label{subsec:nodes}
Each node $n$ in our search algorithm corresponds to a sequence of joint-space actions $\sigma(n) = \{\bolddelta_1, \ldots, \bolddelta_k \}$
and each edge corresponds to one such joint-space action $\delta_i$.
Each node is uniquely  represented by pair $(\mbf{q}^n, \boldk^n)$
 where 
$\mbf{q}^n = \qinit + \delta_1 +  \ldots + \delta_k$ is a joint-space vector 
and~$\boldk^n$ is a vector corresponding the individual curvatures of the robot's sub-segments.
The curvatures~$\boldk^n$ resulting from applying $\sigma(n)$ to $\qinit$ can be obtained by solving the forward kinematics model defined in Eq.~\eqref{eqn:optimisation_problem}. 
The curvatures along with the segment length from the joint space vector can be used to reconstruct the shape of the robot using Eq.~\eqref{eqn:shape_X}.

\subsection{Node extension}
\label{subsection:extnsion}

Our joint-space actions, $\calA := {(0, \delta \lten), (\delta \lseg, \delta \lten)}$, involve either adjusting the robot's tendon by $\delta \lten$ or inserting its base by $\delta \lseg$ while simultaneously adjusting the tendon by $\delta \lten$. 
For a node $n = (\mbf{q}^n, \boldk^n)$ and a joint-space action $\delta \in \calA$, we define the operation of \emph{extending} $n$ by $\delta$ as creating a node $n' = (\mbf{q}^{n'} = \mbf{q}^{n} + \delta, \boldk^{n'})$, where $\boldk^{n'}$ is computed using the optimization in Eq.~\eqref{eqn:optimisation_problem} with $\boldk^{n}$ as the initial guess. 
We say that this extension is \emph{valid} if $\delta$ is a valid joint-space action when applied to $\mbf{q}^n$.

\subsection{Duplicate detection}
\label{subseciton:duplciate}
To avoid expanding similar, yet non-identical states, existing search-based approaches over continuous search spaces (as is our setting) group states into  equivalence classes~\cite{barraquand1993nonholonomic}.
This is analogous to how standard search algorithms (over finite discrete graphs) reduce search effort by identifying previously-visited states, a process known as \emph{duplicate detection}~\cite{dow2009duplicate}.
In our case, two nodes~$n = (\mbf{q}^{n}, \boldk^{n})$
and~$n' = (\mbf{q}^{n'}, \boldk^{n'})$
are considered duplicates if 
(i)~$\mbf{q}^{n} = \mbf{q}^{n'}$
and
(ii)~$\vert\vert\mbf{p}\left({\sigma(n)}\right) - \mbf{p}\left({\sigma(n')}\right) \vert\vert_2 \leq d_{\rm {sim}}$,
where $d_{\rm {sim}} >0$ is some threshold parameter. 
That is, condition (ii) uses the distance between the two tip positions in order to detect duplicates.

\subsection{Heuristic estimate of nodes}
\label{subsection:heuristic}
   
In heuristic-based search, heuristics are used to speed up the search by estimating the cost to reach the goal.
In our case this means estimating the length of a path that will allow the \ee of the TDCR to reach a target pose. 
In this letter, we propose a heuristic that considers the robot's limited steering properties around the known obstacles while avoiding the need to run the more computationally-complex forward kinematic model.

To achieve this, we suggest a simplified contact model for heuristic computation, which will employ constant-curvature (CC) arcs.
Such arcs have been widely used to model load-free configurations~\cite{Webster2010}. 
Specifically, here we model the robot's motion via a sequence of paths having a constant curvature. The robot can only change its curvature along the backbone when in contact with  an obstacle. Specifically, we assume that the shape, and consequently the \ee's trajectory to the goal is a series of mutually tangent CC arcs, with each arc beginning and ending on the surface of an obstacle to reflect the approximated contact mechanics. Therefore, in the absence of any contacts this trajectory would be a single CC arc. 
While CC curves do not accurately reflect the robot shape as there is curvature variation due to contact forces, our work posits that they could be used to provide a sufficient estimate for the cost-to-go as they reflect the TDCR's limited steering capabilities. 

Therefore, our heuristic~$h(n) : \text{SE}(2) \rightarrow \R$ 
takes as input the position and orientation of the robot's \ee at a given node $n$
and estimates the length of the path to reach the goal.
To avoid calculating~$h(n)$ for every prospective node, we precompute~$h(n)$ by performing a backward search from the goal.
We discretize $\text{SE}(2)$ into cells, where each cell stores the length of the shortest path consisting of CC arcs from the goal to it. Using the precomputed values associated with  each  grid cell, the heuristic value for each node can be calculated during the search, as detailed in \ref{subsubsetion:node_heuristic}. 

\subsubsection{Overview}

\begin{figure*}[th!]

\begin{center}
    \begin{overpic}[width=2\columnwidth, grid =False]{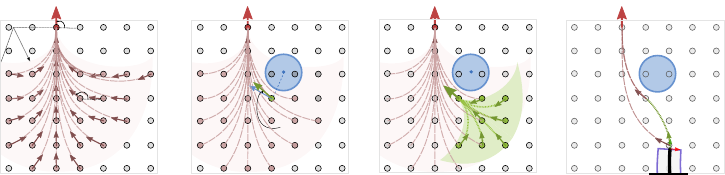}
    \put(2,20.5){\rotatebox{-60}{$\frac{1}{\kappa_{\mrm{max}}}$}}
    \put(12,12){$\angleX(a)$}
    \put(9,22){$\mbf{p}_{\mrm{goal}},\angleX_{\mrm{goal}}$}
    \put(35,22){$\mbf{p}_{\mrm{goal}},\angleX_{\mrm{goal}}$}
    \put(61,22){$\mbf{p}_{\mrm{goal}},\angleX_{\mrm{goal}}$}
    \put(87,22){$\mbf{p}_{\mrm{goal}},\angleX_{\mrm{goal}}$}
    \put(90,11.5){cell $c$}
    \put(85,5){\textcolor{Mahogany}{cell $b$}}
    \put(93,5){\textcolor{OliveGreen}{cell $a$}}
     \put(38,11.5){cell $c$}
     \put(63,11.5){($\mbf{p}_c,\angleX_c$)}
    \put(39,6){$\omega < \omega_{\mrm{contact}}$}
    \put(1,23){(a)}
    \put(26,23){(b)}
    \put(52,23){(c)}
    \put(79,23){(d)}
    \end{overpic}
    \caption{Diagrammatic representation of the heuristic calculation. (a) \steponeinput{$\pose{goal}$} evaluates cell poses that contain CC arcs (red) to reach a goal pose.
    Note that from each cell pose, only one arc reaches the goal.
    (b)
    Having obstacles (blue) invalidates some of the CC arcs. Running \steptwo for each cell pose identified in the previous step identifies contact cells (green circles). 
    (c)
    Using one of the contact cells, cell $c$,  \steponeinput{$\pose{c}$} is run, identifying new CC arcs that reach its pose (green)
    (d)
    After the above steps, we can see
    (i)~positions from which arcs with several orientations were computed (e.g., cell~$b$ and cell~$a$)
    and
    (ii)~For the illustrated robot, the proposed heuristic informs the planner that based on its current end-effector orientation it can reach the goal via the CC arc connecting cell $a$ and $c$, and consequently $c$ to the goal.
    }
     \label{fig:heuristic}
\end{center}
\end{figure*}

In order to precompute the cost of paths of CC arcs from each cell to the goal, we perform a backward search from the goal. We first find the set of all CC arcs whose end point  orientation corresponds to the goal pose (\figref{fig:heuristic}(a)). From this set, we identify cells that could be potential contact points (\figref{fig:heuristic}(b)). From these contact cells, we perform the same search as before for CC arcs, but use the pose of these contact cells to populate consequent CC arcs (\figref{fig:heuristic}(c)). The process is repeated for all possible contact cells. 
Finally, we have a set of CC arcs from each cells that can reach the goal (\figref{fig:heuristic}(d)). 

There are two main components to precomputing the heuristic: (1)~\stepone: Determining the set of all cells that can reach a given target pose with a \emph{valid}  CC arc (to be defined shortly), (2)~\steptwo : Identifies whether an input cell is a possible contact cell. 
This function is evaluated for each cell found in the above set. 
For the case where there are no obstacles, the heuristic computation only involves running step (1) with the goal pose as input.

\subsubsection{Implementation}
\label{subsec:implementation}
As mentioned, our heuristic will rely on a discretization of $\text{SE}(2)$ into cells.
Let this grid be denoted by $\mathtt{C}$, with each cell $c \in \mathtt{C}$ of size $\Delta \times \Delta \times \alpha$ where $\Delta \in \R$ and $\alpha = 2\pi /s$ for some $s \in \N^+$. 
Thus, we introduce  mappings between cells and the pose they represent.
Specifically,
let $\texttt{CELL}$ 
be a mapping taking a pose $\left( \mbf{p}, \angleX \right) \in \text{SE}(2)$ and returning the cell containing this pose.
Similarly, 
let \texttt{POSITION}, \texttt{ORIENTATION} and \texttt{POSE} 
be mappings taking a cell $c \in \mathtt{C}$ and returning the position, orientation and pose at the center of $c$, respectively. The heuristic value of every cell is initially set to $\infty$.

\noindent
\textbf{(1)---\stepone:}
For any target pose $\pose{}\in\text{SE}(2)$, there's a unique CC arc starting from $\position\in\R^2$ and reaching this target pose.
%
Let $a=\texttt{CC-ARC}_{\pose{}} (\mbf{u})$ be this arc. 
Its curvature, and length are denoted by~$\kappa(a)$ and $\ell(a)$, respectively. 
Similarly, its orientation at $\mbf{u}$ w.r.t. the global $x$-axis is denoted by~$\angleX(a)$.
We say that such an arc is \emph{valid} for some curvature bound~$\kappa_{\rm{max}}$  if 
(i)~$\kappa(a) < \kappa_{\rm{max}}$,
(ii)~$\theta(a) < \theta_{\rm{max}}$,
and
(iii)~$a$ does not intersect any obstacle.

Given a position $\position \in \R^2$, let $C_{\position}$ be the set of all cells that have the same position, but different orientations.
Namely,
$$
C_{\position} := \{ c~\vert~\exists \angleX \in \text{SO}(2)~\text{s.t.}~\texttt{CELL}(\position, \angleX) = c\}.
$$

Following the discussion above, 
given some target pose~$\pose{}$, 
if we compute a \emph{valid} arc $a$,
only one cell~$c_a$ in $C_\mbf{q}$ contains this arc's orientation~$\angleX(a)$, i.e. 
$c_a = \texttt{CELL}(\position, \angleX(a))$.
We set the heuristic value $h_a$  of $c_a$ to be the arc's length, i.e. $h_a := \ell(a)$.
We use a breadth-first search (outlined in Alg.~\ref{alg:step1}) to identify the set, $S_{\pose{}}$, of all cells with \emph{valid} arcs reaching $\left(\mbf{p},\angleX\right)$. 
This entire process is equivalent to identifying the workspace of a single CC segment as shown in \figref{fig:heuristic}(a).

\begin{algorithm}[t]
 \caption{Populating the CC arcs for a given pose}
 \begin{algorithmic}[1]
 \renewcommand{\algorithmicrequire}{\textbf{Input:}}
 \renewcommand{\algorithmicensure}{\textbf{Output:}}
 \Require 
  
    Target pose~$\left( \mbf{p}, \angleX \right)$
\Ensure
    $S_{\mbf{p},\angleX}$
\Function{HeuristicArc}{$a$}
     \If {
        $\kappa(a) > \kappa_{\rm{max}}$ 
            \text{or}\\
         $\theta(a) > \theta_{\rm{max}}$ 
            \text{or}\\
        \quad \quad  \quad 
        $\texttt{InCollision}(a)$
        }
        \State \textbf{return} $\infty$
     \EndIf

     \State \textbf{return} $\ell(a)$
\EndFunction
\Function{PopulateCCArcs}{$\mbf{p}$, $\angleX$}
      \State OPEN \la $(\mbf{p})$
      \State $c$ \la \texttt{CELL}$\pose{}$
      \State $S_{\pose{}}$.insert$(c)$
      \While{not OPEN.empty()    }
          \State $\mbf{u}$ \la OPEN.extract()
            \State $a$ \la $\texttt{CC-ARC}_{\left( \mbf{p}, \angleX \right)}(\mbf{u})$
            \State $c_a$ \la \texttt{CELL}($\mbf{u},\angleX(a))$
         \State $h_a$ \la \Call{HeuristicArc}{a}
         \If {$h_a$ $< \infty$}
           \State $h(c_a)$ \la $\min\{h(c_a), h_a + h(c)\}$
            \State $S_{\pose{}}$.insert($c_a$)
            \For{$c'$ in \texttt{PositionNeighbor}($c_a$)}
                \State OPEN.insert(\texttt{POSITION}($c'$))
            \EndFor
        \EndIf
     \EndWhile
     \State \textbf{return} $S_{\pose{}}$
\EndFunction
 \end{algorithmic} 
  \label{alg:step1}
\end{algorithm}

\noindent
\textbf{(2)---\steptwo:}

We say that a cell $c' \neq c$ is a \emph{position neighbor} of $c$ if 
\begin{equation*}
\begin{split}
    &
    \vert
        \texttt{POSITION} (c).x - \texttt{POSITION} (c').x
    \vert
    \leq \Delta \quad \text{and}\\
    &
    \vert
        \texttt{POSITION} (c).y - \texttt{POSITION} (c').y
    \vert
    \leq \Delta.   
\end{split}
\end{equation*}
A cell $c$ is said to be  a \emph{contact cell} if 
(i) there exists a position neighbor $c'$ of $c$ whose position $\texttt{POSITION}(c')$ lies within some obstacle $j$, i.e., $\exists \calO_j\in \calW_{\mathrm{obs}} \text{ s.t. } \calO_j \cap c \neq\emptyset$
and if 
(ii) the orientation of the cell differs from the local tangent of $\calO_j$  by at most~$\omega_{\mrm{contact}}$ i.e., $\vert\texttt{ORIENTATION}(c)-\angleX_{f_j}\left(\texttt{POSITION}(c')\right)\vert < \omega_\mrm{contact}$.
Here, $\omega_{\mrm{contact}}>0$ is some user-provided threshold.

\subsubsection{Algorithm}
To recall, we start by settling the initial heuristic value for all cells to $\infty$. We start by calculating the set of all cells $S_{\pose{goal}}$ that have \emph{valid} CC arcs directed towards the desired goal pose $\pose{goal}$. We do so by calling \steponeinput{$\pose{goal}$} (detailed in Alg.~\ref{alg:step1}) that updates the heuristic value for each cell to be the length of the calculated arc.

We then repeat the  following process (summarized in Alg.~\ref{alg:heuristic}) :
For each $c$ in the priority queue OPEN (extracted in a first-in first-out manner) such that 
$c$ is a contact cell
and
$h(c) < \infty$,
we repeat the previous step by calling \steponeinput{$\left(\texttt{POSE}(c)\right)$} to update the heuristic value of each cell that can reach $c$. If $c_a$ is one such cell that can reach $c$ with a valid arc $a$, we set its heuristic value as
$h(c_a) = \min\{h(c_a), h_a + h(c)\}$

\subsubsection{Returning the heuristic value for a node}
\label{subsubsetion:node_heuristic}
Once the heuristic has been precomputed for grid cells using Alg.~\ref{alg:heuristic}, it can be used to return the heuristic value for a given node. Given a node $n$, 
we compute the end effector $\pose{n}$ of the robot at $n$.
We then compute the cell $c_n$ where this pose lies (i.e., $c_n = \texttt{CELL}(\pose{n})$.
As described earlier, the CC assumption is only an approximation for the contact mechanics and does not precisely reflect contact mechanics. 
Thus, we introduce a final postprocessing smoothing step to account for orientation errors.
Specifically, to set the heuristic value for $c_n$ we also consider the values of it's neighboring cells~$c_n^+$ and~$c_n^-$ which share the same position as $c$ but differ in the orientation by a value of $\pm \alpha$.
Namely,
\begin{equation*}
\begin{split}
    &
     c_n^+ := \texttt{CELL}(
        \mbf{p}_n, \angleX_n + \alpha
    ) \quad \text{and}\\
    &
     c_n^- := \texttt{CELL}(
        \mbf{p}_n, \angleX_n - \alpha
    ).
\end{split}
\end{equation*}
Our final heuristic value for the node $n$ will be
\begin{equation}
    h(n) = 
    \min \{ h(c_n), h(c_n^+), h(c_n^-)\}.
\end{equation}

\begin{algorithm}[t]
 \caption{Algorithm for computing the heuristic}
 \begin{algorithmic}[1]
 \renewcommand{\algorithmicrequire}{\textbf{Input:}}
 \renewcommand{\algorithmicensure}{\textbf{Output:}}
 \Require 
    Target pose~$\left( \mbf{p}_{\mrm{goal}}, \angleX_{\mrm{goal}} \right)$
\Ensure
    $h(c)~\forall c \in \mathtt{C} $
\Function{ComputeHeuristic}{$ \mbf{p}_{\mrm{goal}}, \angleX_{\mrm{goal}} $}
     \State OPEN \la $\left( \mbf{p}_{\mrm{goal}}, \angleX_{\mrm{goal}} \right)$
     \State $h(c)$ \la $\infty$~$\forall c \in \mathtt{C}$
     \State $h(\texttt{CELL}(\pose{goal})$ \la $0$
     \While{not OPEN.empty()    }
      \State $\pose{}$ \la OPEN.extract()
      \State $S_{\pose{}}$ \la \Call{PopulateCCArcs}{$\mbf{p},\angleX$}
      
      \For{$c'$ in $S_{\pose{}}$}

          \If{\texttt{IdentifyContactCell}($c'$)}
          \State OPEN.insert(\texttt{POSE}($c'$))
          \EndIf

     \EndFor
     \EndWhile
     \For {$c$ in $\mathtt{C}$}
     \State  $h(c)$ = 
    $\min \{ h(c), h(c^+), h(c^-)\}$
    \EndFor
     \State \textbf{return} $h$
     
\EndFunction
 \end{algorithmic} 
  \label{alg:heuristic}
\end{algorithm}

\subsection{Putting it all together---Search algorithm}

In this work, we implement a greedy best-first search algorithm~\cite{russell2010artificial}. 
Specifically, we make use of a priority queue \open that orders search nodes (Sec.~\ref{subsec:nodes}) according to the 
the node's heuristic value (Sec.~\ref{subsubsetion:node_heuristic}).

The algorithm starts by initializing \open with a node corresponding to the initial configuration $\qinit$ and runs in iterations.
Each iteration, the highest-priority node $n$ is extracted from \open, and  is \emph{extended} (Sec.~\ref{subsection:extnsion}) by applying the set of actions $\calA$ to its joint space values $\mbf{q}^n$. 
For each successor node $n'$, we compute its corresponding joint space values using our forward kinematic model (Sec.~\ref{section:model}).
If $n'$ is valid and not a duplicate (Sec.~\ref{subseciton:duplciate}), it is added to \open. 
Finally, $n$ is added to a so-called \closed set to avoid re-expanding previously expanded nodes.

This process continues until a node that  satisfies  the end-effector positional constraints specified in Problem~\ref{prob:mp}, at which point the search is concluded successfully or until a maximal number of search iterations~$N_{\mrm{max}}$ is reached at which point the search is classified as unsuccessful.

\section{Empirical evaluation}

We evaluate our contact-based motion planner $\mathcal{M}_{\text{CAN}}$ with our heuristic in simulation. 
Some sample motion plans and their implementation on a motorized prototype~\cite{grassmann2023open} are in the supplementary video.\footnote{Can also be found at \href{https://youtu.be/da3eYGwzxts}{https://youtu.be/da3eYGwzxts}}

\subsubsection{Kinematic model \& planner parameters}
The optimization problem in Eq.~\eqref{eqn:optimisation_problem} is solved using an interior-point algorithm via MATLAB's \textit{fmincon}. 
We consider a robot of radius \SI{6}{mm} with a maximum length of \SI{250}{mm}, represented by $m=30$ arcs. 
The planner parameters are listed in Table~\ref{table:hyperparameters_}.

\begin{table}[h!]
\centering
\caption{Parameters for the heuristic calculation and motion planner.}
\begin{tabular}{|c|c|c|c|}
\hline
\multicolumn{2}{|c|}{Heuristic} & \multicolumn{2}{c|}{Motion Planner} \\
\hline
Parameter & Value & Parameter & Value \\
\hline
$\Delta$& \SI{0.001}{m} & $\epsilon$ & \SI{0.01}{m} \\
$\alpha$& \SI{45}{\degree} & $\omega$ & \SI{15}{\degree} \\
$\kappa_{\mrm{max}}$ & $250$ $m^{-1}$ & $\delta \lseg$ & \SI{0.001}{m} \\
$\theta_{\mrm{max}}$ & $\SI{270}{\degree}$ $m^{-1}$ & $\delta \lten$ &  \SI{0.001}{m} \\
$\omega_{\mrm{contact}}$& \SI{2.815}{\degree} &  $N_{\mrm{max}}$ & 7000 \\
\hline
\end{tabular}
\label{table:hyperparameters_}
\end{table}

\subsubsection{Environments}
We consider three workspaces,~$\calW_1$,$\calW_2$, and $\calW_3$ (see \figref{fig:analysis}). 
$\calW_1$ has five equally sized convex circular obstacles while $\calW_2$ features six circular obstacles with random centers and radii that overlap, creating non-convex shapes.
Finally, $\calW_3$ resembles a 2D cross-section of the interior of a turbine engine. CAN's potential use in such cluttered environments guides our future work (refer to Sec.~\ref{sec:future}).
Specifically, 
the non-convex turbine blade shapes are approximated by four circular obstacles each.

Recall that we compute a  heuristic for the robot's end effector (Sec.~\ref{subsection:heuristic}) which can be modelled as a line segment translating and rotating in the plane.
Thus, to ensure that we only consider collision-free end effector placements, 
heuristic computation is done after inflating obstacles by an amount equal to the robot's width~\cite{halperin2017algorithmic}.

\subsubsection{Queries}
We generated for each workspace a set of valid queries as follows:
We start by setting the initial configuration to~$\qinit=(1,1)mm$. We then run a breadth-first search (BFS) by varying the tendon and segment length by \SI{1}{mm}.
This creates a set of valid configurations with corresponding end-effector poses for which a solution to Problem~\ref{prob:mp} is guaranteed to exist.
For each workspace, we sample 175 such end-effector poses at random. 

\subsubsection{Baselines.}
We compare our planner~$\mathcal{M}_{\text{CAN}}$ with two baselines.
The first baseline, which we denote by $\mathcal{M}_{\text{contactless}}$, is a planner that avoids contacts with the obstacles. 
Specifically, in the  kinematic model, contactless configurations have constant curvatures.
Thus, $\mathcal{M}_{\text{contactless}}$ performs inverse kinematics~\cite{Webster2010} for points satisfying the position tolerance and checks if the obtained solution satisfies the angular tolerance as well and does not intersect with obstacles. 

The second baseline, which we denote by $\mathcal{M}_{\text{CAN-simple}_h}$ is identical to our planner but uses a simple commonly-used heuristic function $h_{\text{simple}}: = h_{xyz}+w\cdot h_{\theta}$~\cite{Cohen2010}.
Here~$h_{xyz}$ represents the least-cost path from the tip to the goal, computed via Dijkstra's algorithm,
$h_{\theta}$ is the absolute angular difference between the end-effector and the target orientation,
and $w$ is a constant used to balance the two, set to be $0.01$.

\subsubsection{Success rate for different planners}
In Table~\ref{table:searchresult-}, we present each planner's success rate in different workspaces. 
Both baselines show significantly lower success rates, underscoring the need for CAN and the proposed heuristic. 
Failures of $\mathcal{M}_{\text{CAN}}$ could be attributed to two factors:
The first is that the heuristic's inexact approximation of contact mechanics may not always guide the robot successfully. 
The second is that, in many cases, if a particular path does not result for a given sequence of contacts, the planner tends to explore nearby nodes, without exploring paths with alternative contacts.
A potential remedy is to discern similar configurations through soft duplicate detection~\cite{DuKSL19,Maray2022}.

\begin{table}[h]
\centering
\caption{Success rate [\%] of finding a solution for the different planning approaches on the three workspaces.}

\begin{tabular}{|c|c|c|c|}
\hline
 & $\calW_1$ & $\calW_2$ & $\calW_3$ \\
\hline
$\mathcal{M}_{\text{CAN}}$ (proposed planner) & 80.57 & 79.43 & 78.86 \\
\hline
$\mathcal{M}_{\text{contactless}}$  (baseline 1) & 16.57 & 25.14 & 26.86 \\
\hline
$\mathcal{M}_{\text{CAN-simple}_h}$ (baseline 2)& 29.71 & 19.42& 28.57 \\
\hline
\end{tabular}
\label{table:searchresult-}
\end{table}

\subsubsection{Comparing the planner with BFS.}
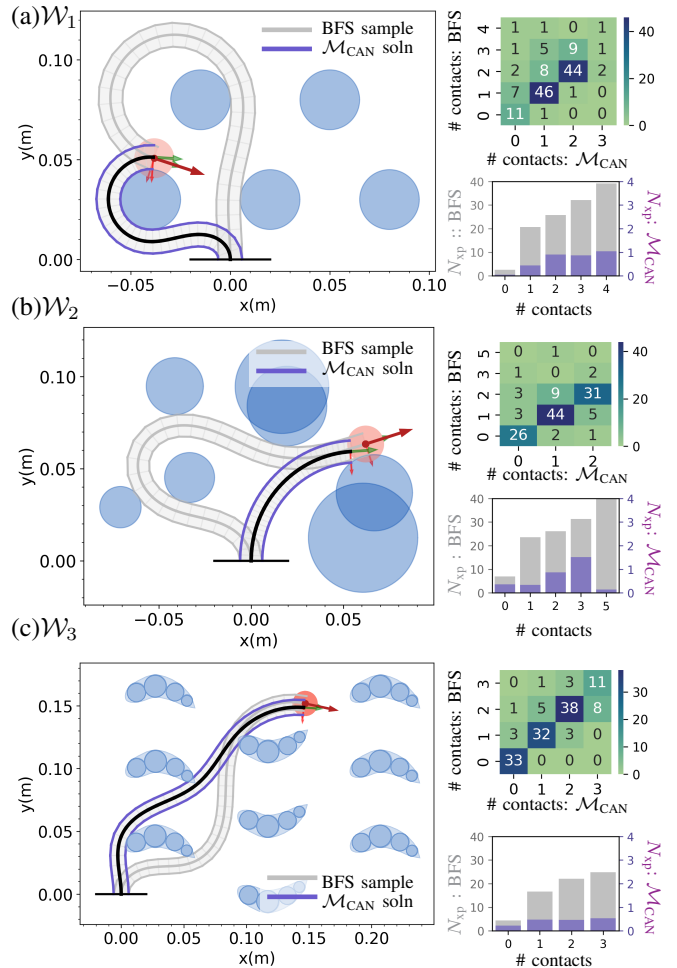
\begin{figure}[t!]
    \begin{overpic}[width=1\columnwidth, grid = False]{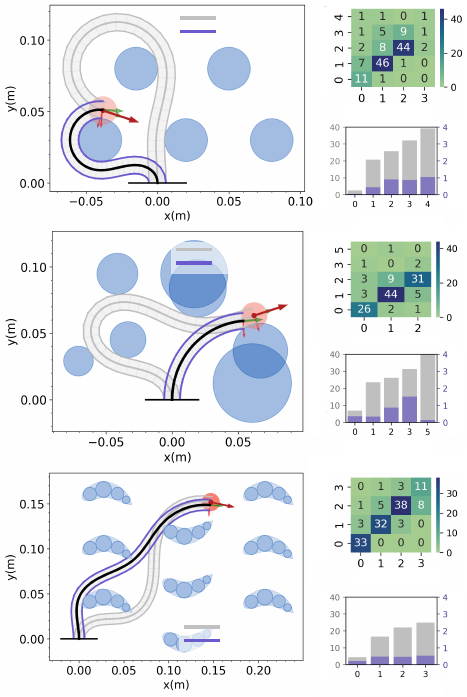}
    \put(0,98){(a)$\calW_1$}
    \put(0,68){(b)$\calW_2$}
    \put(0,35){(c)$\calW_3$}
    \put(32,97){\scriptsize{BFS sample}}
    \put(32,95){\scriptsize{$\mathcal{M}_{\text{CAN}}$ soln}}
    \put(32,64){\scriptsize{BFS sample}}
    \put(32,62){\scriptsize{$\mathcal{M}_{\text{CAN}}$ soln}}
    \put(32,9){\scriptsize{BFS sample}}
    \put(32,7){\scriptsize{$\mathcal{M}_{\text{CAN}}$ soln}}
    \put(45,87){\rotatebox{90}{\fontsize{7pt}{8pt}\selectfont \# contacts: BFS}}
    \put(49,83.5){\rotatebox{0}{\fontsize{7pt}{8pt}\selectfont \# contacts: \mcan}}
    \put(45,72){\rotatebox{90}{\fontsize{7pt}{8pt}\selectfont \textcolor{Gray}{$N_{\text{xp}}:$: BFS}}}
    \put(65,81){\rotatebox{-90}{\fontsize{7pt}{8pt}\selectfont \textcolor{Plum}{$N_{\text{xp}}$: \mcan}}}
    \put(52,68){\rotatebox{0}{\fontsize{7pt}{8pt}\selectfont \# contacts}}

    \put(45,52){\rotatebox{90}{\fontsize{7pt}{8pt}\selectfont \# contacts: BFS}}
    \put(49,51){\rotatebox{0}{\fontsize{7pt}{8pt}\selectfont \# contacts: \mcan}}
    \put(45,39){\rotatebox{90}{\fontsize{7pt}{8pt}\selectfont \textcolor{Gray}{$N_{\text{xp}}:$ BFS}}}
    \put(65,49){\rotatebox{-90}{\fontsize{7pt}{8pt}\selectfont \textcolor{Plum}{$N_{\text{xp}}$: \mcan}}}
    \put(52,35){\rotatebox{0}{\fontsize{7pt}{8pt}\selectfont \# contacts}}

    \put(45,19){\rotatebox{90}{\fontsize{7pt}{8pt}\selectfont \# contacts: BFS}}
    \put(49,17){\rotatebox{0}{\fontsize{7pt}{8pt}\selectfont \# contacts: \mcan}}
    \put(45,4){\rotatebox{90}{\fontsize{7pt}{8pt}\selectfont \textcolor{Gray}{$N_{\text{xp}}:$ BFS}}}
    \put(65,14){\rotatebox{-90}{\fontsize{7pt}{8pt}\selectfont \textcolor{Plum}{$N_{\text{xp}}$: \mcan}}}
    \put(52,1){\rotatebox{0}{\fontsize{7pt}{8pt}\selectfont \# contacts}}
    
    \end{overpic}
    \caption{Comparison of our motion planner~$\mathcal{M}_{\text{CAN}}$ with the  BFS planner used to generate the queries across three workspaces. 
    For each plot we report
    (Top right) a heatmap showing the number of successful solutions for both BFS (y-axis) and $\mathcal{M}_{\text{CAN}}$ (x-axis), with numerical counts displayed within each cell.
    (Bottom right) $N_{\text{xp}}$, the average nodes expanded  ($\times1e3$ for $\mathcal{M}_{\text{CAN}}$ (purple) and BFS (grey)) with increasing number of contacts. Please note that the y-axis scale differs between the two planners.
    }
    \label{fig:analysis}
\end{figure}

We compared our planner $\mathcal{M}_{\text{CAN}}$ with the BFS planner used to generate queries, focusing on the number of contacts and the average nodes expanded (see Fig.~\ref{fig:analysis}). 
The heatmap analysis indicates that most solutions share a similar number of contacts, with a tendency for $\mathcal{M}_{\text{CAN}}$ to identify solutions with fewer contacts if possible. 
Notably, for $\calW_2$ with five contacts, $\mathcal{M}_{\text{CAN}}$ finds a simpler solution, reflected in a lower value in the barplot.
The barplot shows that the average number of nodes expanded by both planners increase with number of contacts, with BFS expanding more than 10 times the nodes expanded by~$\mathcal{M}_{\text{CAN}}$.

\section{Conclusion \& future work}
\label{sec:future}
In this letter, we present a motion planner coupled with a novel heuristic that computes plans to reach a desired pose.
Achieving a success rate of roughly $80\%$ over 525 target poses amidst convex and non-convex obstacles.
This efficiency stems from the proposed heuristic that enables the planner to examine just a fraction of the entire search space.

We consider several directions for future work which we briefly elaborate on:

\subsubsection{Providing guarantees on the quality of the solution.}

The heuristic we propose in Section~\ref{subsection:heuristic} is not guaranteed to be admissible, meaning it does not provide a lower bound on the cost-to-go.
Consequently, there is not guarantee on the quality of the obtained solution by our planner. 
A potential solution to this limitation is employing multi-heuristic A* algorithms, which effectively incorporates both admissible and inadmissible heuristics in a structured manner~\cite{AineSNHL16,IslamSL18}. 

\subsubsection{Extending the planner to 3D environments.}
To extend $\mathcal{M}_{\text{CAN}}$ for use in 3D environments, both the model and the heuristic require modifications. 
The design of the proposed motion planner is model-agnostic, allowing the potential use of any model in literature. 
The heuristic would require voxelizing the spaces~$\R^3$ and $SO(3)$. Additionally, a challenge would be to represent the potential twists in the robot's structure due to contact interactions.
To address this, the heuristic could be computed using arcs of uniform curvature along the $x$ and $y$-axes, combined with a constant rate of twist around the $z$-axis. 

An additional challenge is the computational effort: The current 2D kinematic model accounts for 99.3\% of the computational effort, and shifting to a 3D model could increase this due to the complexity of TDCR modeling. 
Using a simplified approach, such as Euler curves~\cite{Rao2022}, to model the backbone under contact forces might offer a feasible alternative.
Another alternative may be to use multiple resolutions during the search~\cite{FuSA21,FuSSA22}.

\subsubsection{Using the planner for inspection of turbine engines.}
While 21 DOF hyperredundant robots have been investigated for turbine blade inspections~\cite{Saleem2021}, TDCRs with CAN offer the advantage of performing the same task with fewer DOFs. TDCRs mitigate the risk of blade damage due to their compliance. With our proposed motion planner targeting specific poses, future work will evolve into search-based inspection planning~\cite{FuKSA23}, focusing on surface inspection by orienting the robot appropriately.

\section*{Acknowledgements}
The authors would like to thank Quentin Peyron for discussions aiding in the conceptualisation of this problem statement, Puspita Triana Dewi and Chloe Pogue for helping with prototyping and experiments.
\bibliographystyle{IEEEtran}
\bibliography{IEEEabrv,survey}
\end{document}